\documentclass{article} 
\usepackage[preprint]{colm2026_conference}

\usepackage{microtype}
\usepackage{hyperref}
\usepackage{url}
\usepackage{booktabs}
\usepackage{graphicx}
\usepackage[htt]{hyphenat}
\usepackage{amsmath}
\usepackage{dutchcal}
\usepackage{booktabs}
\usepackage{tcolorbox}
\usepackage{multirow}
\usepackage{tabularx}
\usepackage[dvipsnames]{xcolor}
\usepackage{xspace}
\usepackage{wrapfig}

\usepackage{lineno}

\definecolor{darkblue}{rgb}{0, 0, 0.5}
\definecolor{onlyenglish}{HTML}{47A184}
\definecolor{majority}{HTML}{D86D44}
\definecolor{oracle}{HTML}{6E80A8}
\definecolor{country}{HTML}{C36BA2}
\definecolor{globallanguage}{HTML}{85B636}
\definecolor{llmselected}{HTML}{DBB614}
\definecolor{lskextractor}{HTML}{C1A274}

\hypersetup{colorlinks=true, citecolor=darkblue, linkcolor=darkblue, urlcolor=darkblue}

\title{Language Specific Knowledge: Do Models Know Better in \textit{X} than in English?}

\author{%
    Ishika Agarwal$^*$, Nimet Beyza Bozdag$^*$, Dilek Hakkani-Tür \\
  Department of Computer Science\\
  University of Illinois, Urbana-Champaign\\
  \texttt{\{ishikaa2, nbozdag2, nisvalp2, dilek\}@illinois.edu} \\
}

\newcommand{\sysn}{\textsc{LSKExtractor}}

\newcommand{\onlyenglish}{\textcolor{onlyenglish}{Only English}\xspace}
\newcommand{\majority}{\textcolor{majority}{Majority}\xspace}
\newcommand{\globallanguage}{\textcolor{globallanguage}{GlobalLanguage}\xspace}
\newcommand{\llmselected}{\textcolor{llmselected}{LLMSelected}\xspace}
\newcommand{\country}{\textcolor{country}{Country}\xspace}
\newcommand{\lskextractor}{\textcolor{lskextractor}{LSKExtractor}\xspace}
\newcommand{\oracle}{\textcolor{oracle}{Oracle}\xspace}

\begin{document}

\ifcolmsubmission
\linenumbers
\fi

\maketitle
\begin{abstract}

Often, multilingual language models are trained with the objective to map semantically similar content (in different languages) in the same latent space. In this paper, we show a nuance in this training objective, and find that \textit{by changing the language of the input query, we can improve the question answering ability of language models}. We make two main contributions. First, we introduce the term \textbf{Language Specific Knowledge (LSK)} to denote queries that are best answered in an ``expert language'' for a given LLM, thereby enhancing its question-answering ability. We introduce the problem of language selection---for some queries, language models can perform better when queried in languages other than English, sometimes even better in low-resource languages---and \textit{the goal is to select the optimal language for the query}. Second, we introduce a variety of simple to strong baselines to empirically motivate the language selection problem (including one of our own methods called \sysn). During our evaluation, we employ three datasets that contain knowledge about both cultural and social behavioral norms. Overall, the results show that principled language selection can improve the performance of a language model, and that the expected question-to-language map is not always intuitive: Gemma models know most about China and Middle East in Spanish; Qwen models know most about authority and responsibility in Arabic and Chinese. Broadly, our research contributes to the open-source\footnote{https://anonymous.4open.science/r/LSKExtractor-272F/} development of language models that are inclusive and more aligned with the cultural and linguistic contexts in which they are deployed.
\end{abstract}

\vspace{-8pt}
\section{Introduction}

\vspace{-8pt}
Language models are trained to understand and generate responses in dozens of languages, and are trained with either monolingual or parallelly translated data \citep{aya_dataset}. Multilingual language models are trained so that two sentences that are semantically similar but in different languages are mapped to the same point in the latent space \citep{xu2025survey, curseofmultilinguality, pfeiffer-etal-2022-lifting, ruder2019survey} (what we coin as the \textbf{"latent language alignment hypothesis"}). This hypothesis applies to sentences in all languages, creating multilingual language models. This hypothesis is supported by current reports on DeepSeek-R1 \citep{deepseekr1} spontaneously switching to Chinese during its chain-of-thought, even when presented with an English query \citep{thoughtology}. However, the same hypothesis has been challenged by works like the Multilingual Trolley Problem \citep{jin2025language} which show that the alignment of multilingual language models to human preferences varies with the language of the input query.

\begin{figure}[t]
    \centering
    \includegraphics[width=0.9\columnwidth]{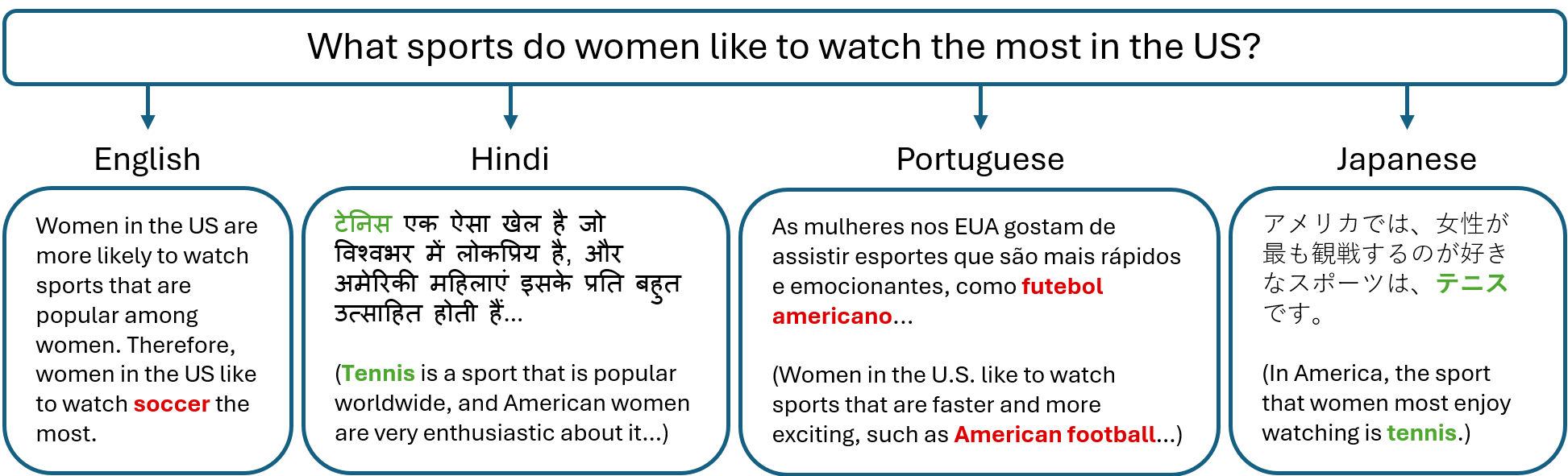}
    \caption{In this toy experiment, we prompt Llama-3.1-8B-Instruct with the same question across multiple languages (shown in English here only for illustration; the actual queries were translated into each respective language). The correct answer is \textbf{tennis}, yet the model produces different outputs depending on the query language. This illustrates what we refer to as Language-Specific Knowledge.}
    \label{fig: motivating_example}
    \vspace{-10pt}
\end{figure}

Figure \ref{fig: motivating_example} presents another case in which the hypothesis of latent language alignment does not hold. In this example, we ask Llama-3.1-8B-Instruct about the sport that American women tend to watch the most in different languages (see the caption for details of this toy experiment). The model produces different answers across languages, with only Hindi and Japanese yielding correct responses. If the languages were truly aligned in the latent space, we would expect the model to produce the same output regardless of the input language. This inconsistency highlights limitations of the latent language alignment hypothesis, which arise from known sources of cross-lingual misalignment such as non-compositionality (the meaning of a phrase cannot be deduced directly from the individual words, i.e., metaphors and idioms \citep{sathe2024language, cheng2024no, clcl}) and non-isomorphism (words lacking direct translations \citep{wu2024representational}). Building on this perspective, we propose another source of misalignment---this time within language models themselves---which we call Language Specific Knowledge.

We define \textbf{Language Specific Knowledge (LSK)} as knowledge that a language model appears to access more readily or represent more accurately when queried/asked to reason in a particular language (the \textit{expert language}). In Figure \ref{fig: motivating_example}, the varying responses across languages are evidence of LSK. Rather than viewing this as a limitation, we argue that such behavior should be leveraged in a more informed and intentional manner, allowing us to guide language models toward languages that may yield more accurate, aligned, and culturally appropriate responses for a given topic.

In this work, our contributions are as follows:
\begin{itemize}
    \itemsep -0.5ex
    \item We formally define \textbf{Language Specific Knowledge (LSK)} and provide intuitive and empirical evidence of its presence in multilingual language models.
    \item We introduce the \textbf{Language Selection Problem} and benchmark a variety of simple to strong baselines that attempt to solve it. We also highlight gaps to inspire future work.
    \item We conduct \textbf{systematic experiments} across multiple state-of-the-art models and three datasets (CultureAtlas \citep{cultureatlas}, BLEnD \citep{blend}, and Social IQa \citep{socialiqa}), using language-specific chain-of-thought reasoning to evaluate the effects of language selection on performance across topics.
    \item We outline that the LSK problem is not a data problem, where models do not contain certain knowledge. This is clear from the near 100\% performance on a method called \oracle. LSK comes from the ineffective strategies to elicit the knowledge from an LLM. Thus, \textbf{we introduce \oracle as a benchmark}, to set the upper-bound of LSK-related knowledge gaps.
\end{itemize}


\vspace{-8pt}
\section{Related Work}
\vspace{-8pt}
Prior work has examined how language influences model reasoning {\citep{schut2025multilingualllmsthinkenglish, zhong2024englishcentricllmslanguagemultilingual, yong2025crosslingualreasoningtesttimescaling}, effects of language on model alignment with human preferences \citep{jin2025language, durmus2024towards}, and cross-linguistic generalization {\citep{chang2022geometrymultilinguallanguagemodel}. \citet{chang2022geometrymultilinguallanguagemodel} investigated how different languages are represented within the XLM-R multilingual model. They found that languages occupy distinct regions in the representational space, though languages with similar distributions can be aligned through mean-shifting. This indicates that semantically equivalent sentences in different languages may not map to the same low-level representations. This insight motivates our study by highlighting the need for language-specific knowledge representations when reasoning or answering questions across linguistic boundaries.

Other works have focused specifically on multilingual reasoning. For instance, \citet{schut2025multilingualllmsthinkenglish} demonstrated that language models tend to default to English during internal reasoning, which can negatively impact downstream task performance, fluency, and fairness. We extended this finding by identifying, for some given topic, the language in which a multilingual language model exhibited greater expertise. Similarly, \citet{zhong2024englishcentricllmslanguagemultilingual} found that models often reason internally in a specific language and exhibit cultural biases aligned with that language when responding to culturally grounded questions. In our work, we aim to boost multilingual reasoning by identifying such LSK and strategically leveraging expert languages where such knowledge is most richly encoded. This complements approaches like \citet{huang2024mindmerger} that merge external multilingual representations to enhance general understanding, \citet{ziabari2025reasoningspectrumaligningllms}, which adapt LLM reasoning between intuitive (System 1) and deliberative (System 2) modes based on task needs, or even \citet{xlingualthoughtprompting} that encourages language models to think in other languages to improve performance. We adapt this as a baseline, called the LLMSelected baseline.

Several works have investigated multilingual reasoning from different perspectives: improving reasoning in low-resource languages \citep{senel2024kardecs}, benchmarking the reasoning abilities of language models across languages \citep{etxaniz2023multilinguallanguagemodelsthink, dynamic_learning, gao2025thinkingmultilinguallyempowerllm}, and enhancing semantic alignment between languages \citep{langbridge}. These efforts primarily aim to strengthen cross-lingual semantic representations to support more consistent reasoning across languages. In a related line of work, \citet{yong2025crosslingualreasoningtesttimescaling} demonstrated that chain-of-thought traces in various languages can be aligned to their English counterparts to facilitate multilingual reasoning. In contrast, we highlight a fundamental limitation of this alignment approach: certain languages encode concepts that do not have direct equivalents in others. This observation underscores the lack of a universal one-to-one mapping across languages \citep{maps_dataset}. \textbf{Rather than enforcing alignment, our work embraces linguistic diversity by leveraging the unique conceptual affordances of each language to enhance reasoning performance.}

Furthermore, language is an important part of model alignment with human preferences. However, prior work has shown that current multilingual models are not well aligned with humans, showing more US and Euro-centric representations rather than multicultural \citep{durmus2024towards, rystrøm2025multilingualmulticulturalevaluating}. Recent studies have shown that languages are indeed proxies for culture \cite{accordingtosagnik}, thus they should be aligned to culturally diverse preferences. However, even when prompted across different languages, they fail to align with these culturally diverse moral preferences \citep{jin2025language}. Our work contributes to alignment by identifying the expert language for specific domains of knowledge and demonstrating how strategically using these languages can elicit responses that better reflect localized, culturally grounded human preferences.

\vspace{-5pt}
\section{The Language Selection Problem}
\vspace{-3pt}
Formally, given a query $Q$ and a set of candidate languages $\mathcal{L}$, the goal of the Language Selection Problem is to select a language $\ell \in \mathcal{L}$ in which to reason about $Q$ in, in order to optimize the language model's performance. To clarify, the selected language, used for chain-of-thought reasoning, can differ from the language of the original question. The final answer is always returned in the original language of $Q$.

\vspace{-5pt}
\paragraph{Practical implementation.} In our setup, we first \textit{translate the queries} from the source language (mostly English) to each candidate language $\ell \in \mathcal{L}$ and perform chain-of-thought reasoning in each. Next, we \textit{categorize the resulting answers} as correct or incorrect. Since our evaluation datasets (CultureAtlas \citep{cultureatlas}, BLEnD \citep{blend}, and Social IQa \citep{socialiqa}) are multiple choice, correctness is determined via answer matching.\footnote{For text-generation tasks with free-response answers, the method can be adapted by deeming a response correct when the model's prediction and the ground-truth answer are (nearly) semantically equivalent.} We then \textit{apply each language selection method} to choose a single language per query and evaluate based on the accuracy of the selected language's response.

\subsection{Methods}
\label{sec: language_selection_methods}

\begin{figure}
    \centering
    \includegraphics[width=0.49\linewidth]{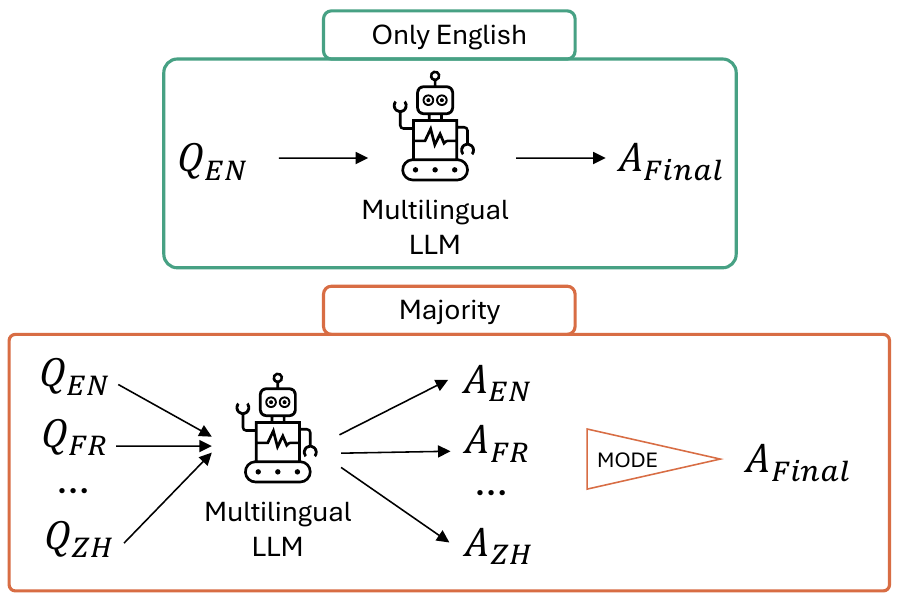}
    \includegraphics[width=0.49\linewidth]{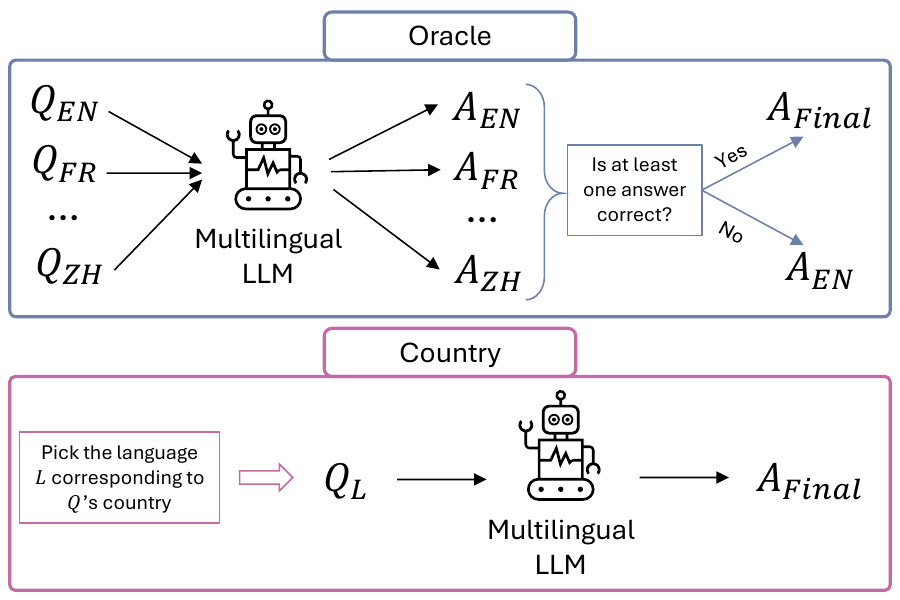}
    \includegraphics[width=0.49\linewidth]{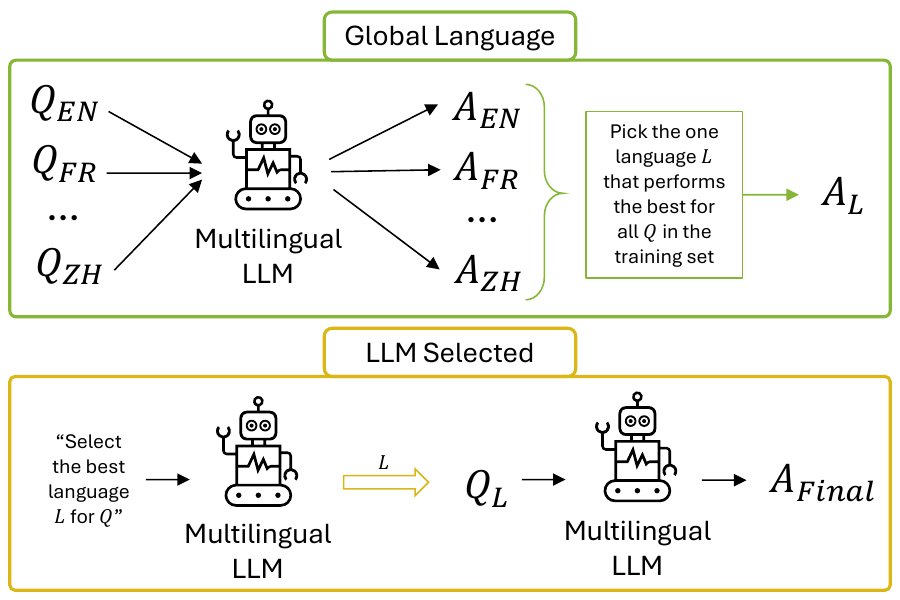}
    \includegraphics[width=0.49\linewidth]{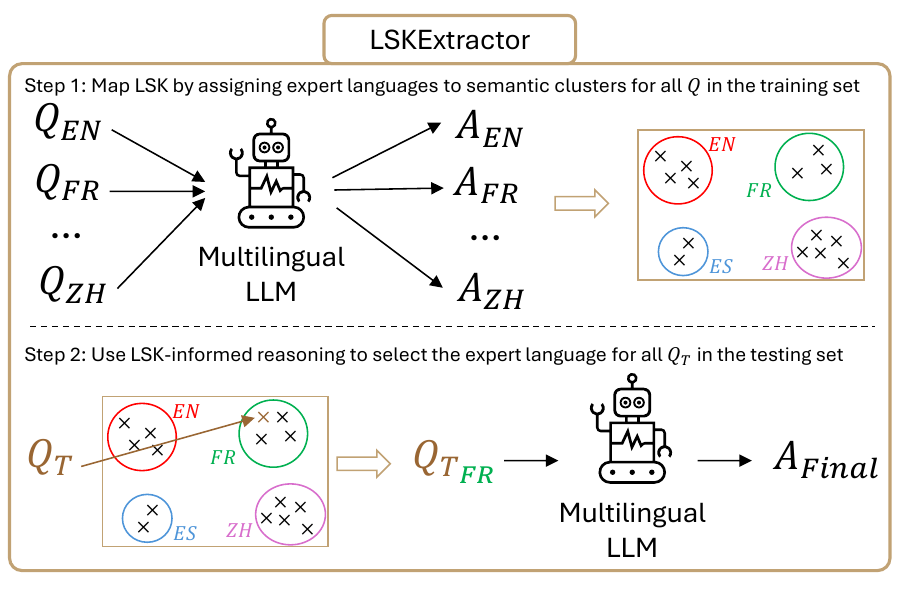}
    \caption{Illustrations of the seven different language selection methods we use in our LSK evaluation. Detailed descriptions of each of them are presented in Section \ref{sec: language_selection_methods}.}
    \label{fig: language_selection_methods}
\end{figure}

Figure \ref{fig: language_selection_methods} illustrates each of the seven language selection methods that we use to motivate the language selection problem. Some of the baselines below use a training and testing split. We elaborate on these details in Section \ref{sec: experiments}. For now, below are detailed descriptions of these language selection methods:

\begin{itemize}
    \item \textbf{\onlyenglish}: This is our \textbf{most simple} baseline. We simply ask the language model to reason about the question in English. There is no language selection in this method. By using this as a baseline, we are able to prove that the language selection problem is a worthwhile problem -- that is, that reasoning in a language other than English can yield measurable improvements.
    
    \item \textbf{\majority}: This is our \textbf{most computationally expensive} method. We perform chain-of-thought reasoning in every candidate language and aggregate answers via majority voting. This serves as one of the upper bounds for the language selection problem, since it leverages information from all languages rather than selecting a single one.
    
    \item \textbf{\globallanguage}: Using a training set, we identify the single language $L$ that achieves the highest overall accuracy and use it for inference on all queries in the testing set. If the language is English, it is equivalent to \onlyenglish. But there is no guarantee that the selected language will be English.
    
    \item \textbf{\llmselected}: We prompt the language model itself to choose the most appropriate reasoning language for a given query, then have it answer the question in that language. The prompt for this is in Appendix \ref{app: model_prompts}, Table \ref{prompt: llm_selected_prompt}.
    
    \item \textbf{\country}: We select the reasoning language based on the country of origin associated with the query. Note that not all datasets include this metadata. Details of the country-to-language mapping are provided in Appendix~\ref{app:country_mapping}.
    
    \item \textbf{\lskextractor}: This is \textbf{a novel method} that we propose in this paper. It relies on the assumption that LSK occurs \textit{semantically} (i.e., models know most about a certain \textit{topic} in a particular language). This is a two-step approach. In Step 1, we embed training queries into a shared semantic space and cluster them based on topical similarity. For each cluster, we determine the \textit{expert language}: the language that yields the highest accuracy across the cluster's queries. In Step 2, during test-time inference, we embed the test query into the same space, identify its nearest cluster, and select the corresponding expert language (e.g., French) to guide the model toward producing a more informed and culturally grounded response.
    
    \item \textbf{\oracle}: This method represents our \textbf{upper bound}, as it measures the performance of the perfect language selector. For each query, we select (in hindsight) the language in which the model answered correctly.
\end{itemize}

\section{Experiments}
\label{sec: experiments}
\subsection{Experimental Setup}
\label{sec: exp_setup}

\begin{wraptable}{r}{0.4\textwidth}
    \centering \small
    \begin{tabular}{c|c}
    \toprule
    Language & Avg. Score \\ \midrule
    Hindi & 3.83 \\
    Spanish & 3.93 \\
    Chinese & 3.70 \\
    Turkish & 3.87 \\
    Portuguese & 3.60 \\
    Arabic & 2.83 \\
    Russian & 2.87 \\
    Korean & 3.63 \\
    Vietnamese & 3.63 \\
    \bottomrule
    \end{tabular}
    \caption{Average scores (out of 4) for translation quality, on a subset of 30 samples, judged by human annotators.}
    \label{tab:human_verification}
\end{wraptable}

\paragraph{Languages.} For our experiments, we set $\mathcal{L}$ to include the following 16 languages: Arabic, Bengali, Chinese, English, French, German, Hindi, Italian, Japanese, Korean, Portuguese, Russian, Spanish, Thai, Turkish, and Vietnamese. It is important to note that $\mathcal{L}$ is treated as a hyperparameter: the methodology selects the best expert language from the available set. Crucially, a model’s multilingual coverage does not need to align with the chosen set of languages, since the language selection problem aims to select the most appropriate language possible. The (multiple choice question-answering) datasets we choose are in English; hence, to ensure the integrity of our experimental design and that models truly reason in the particular language selected by methods in Figure \ref{fig: language_selection_methods}, we translate the instructions and inputs from the datasets (below) into the 16 languages we outlined above. We use OpenAI's GPT-4o-mini to translate the queries (instruction + input + answer choices). Our prompt is outlined in Figure \ref{prompt: translation}.

In order to check GPT's translation quality, we ask humans to verify the translation quality. On a subset of 10 samples per dataset (totaling 30 samples), we asked participants to rate the quality of the translation from 1 (nonsense translation) to 4 (perfect translation). The average rating, broken down by language, is in Table \ref{tab:human_verification}.


\paragraph{Datasets.} We hypothesize that language-specific knowledge may manifest itself in culture, societal norms, and common sense reasoning. Math, coding, and logic are examples of domains that we expect to have little LSK, which is why we do not evaluate on those domains. Hence, we select three datasets that reflect these properties:

\begin{itemize}
\itemsep -0.5ex
    \item \textbf{CultureAtlas} \citep{cultureatlas}: a dataset consisting of cultural norms (e.g., ``During the Chinese New Year, in Southern China, red envelopes are typically given by the married to the unmarried[...]''), labeled as either True or False. To create a more challenging task, we reformat the dataset into multiple-choice questions (MCQs) with four answer options: one true claim and three false claims about the same country. Further details are provided in Appendix~\ref{app:cultureatlas_reformatting}.
    \item \textbf{BLEnD} \citep{blend}: a multiple-choice question answering dataset where the input is a societal norm (e.g., "What is the common dress code for school teachers in Azerbaijan?") and four answer choices (e.g., "A. apron, B. black formal suit, C. uniform, D. shirt"). The output is one of the selected answer choices.
    \item \textbf{Social IQa} \citep{socialiqa}: a multiple-choice common sense reasoning dataset where the input contains some context (e.g., "Sydney walked past a homeless woman asking for change but did not have any money [...] Sydney felt bad".), a question (e.g., "How would you describe Sydney?"), and three answer choices (e.g., "A. sympathetic, B. like a person who was unable to help, C. incredulous"). The output is one of the answer choices.
\end{itemize}

The \majority, \globallanguage, and \lskextractor methods use a separate training and testing split. Hence, we use 8k instances for training and 2k for testing on BLEnD and Social IQa datasets. For CultureAtlas, due to reformatting, we use 5k instances for training and 1.5k for testing. Since all datasets are framed as question-answering tasks, \textbf{we measure and report performance with classification accuracy} ((\# of True Positive + \# of True Negative) / \# of All Predictions).

\paragraph{Models.} For our evaluation, we use a variety of model sizes from a variety of families: Google's \texttt{gemma-3-1b-it} and \texttt{gemma-3-12b-it} \citep{gemmateam2024gemmaopenmodelsbased}, Meta's \texttt{Llama-3.2-1B-Instruct}, \texttt{Llama-3.2-3B-Instruct}, and \texttt{Llama-3.1-8B-Instruct} \citep{grattafiori2024llama3herdmodels}, Qwen's \texttt{Qwen3-0.6B}, \texttt{Qwen3-8B}, and \texttt{Qwen3-14B} \citep{yang2025qwen3technicalreport}, and CohereLab's \texttt{aya-23-8B} \citep{dang2024ayaexpansecombiningresearch}. We use instruction-tuned versions because those models are trained to handle multilingual inputs. To ensure they generate outputs in the correct LSK language, we run a language classification on the output (details are in Appendix \ref{app: output_analysis}). Finally, since \lskextractor requires an embedding model, we use the \texttt{Qwen3-0.6B-Embedding} model.

\subsection{Results and Analysis}

The results are in Figure \ref{fig: main_wr}. Below, we analyze the results and offer our insights into why the language selection problem is an important one to solve, and where the gaps are in current methods.

\begin{figure}[h]
    \centering
    \makebox[\textwidth][c]{\includegraphics[width=1\columnwidth]{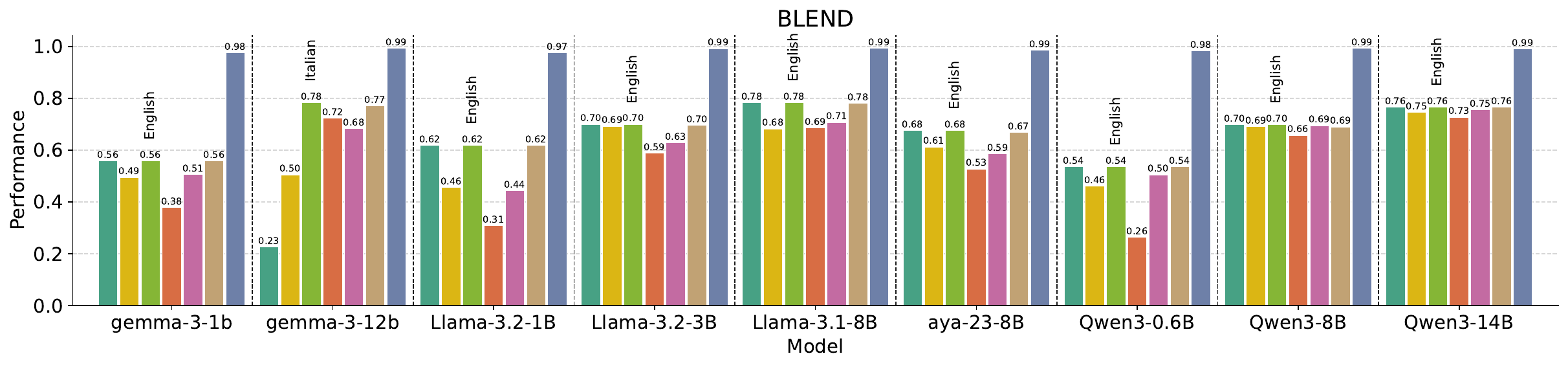}}
     \makebox[\textwidth][c]{\includegraphics[width=1\columnwidth]{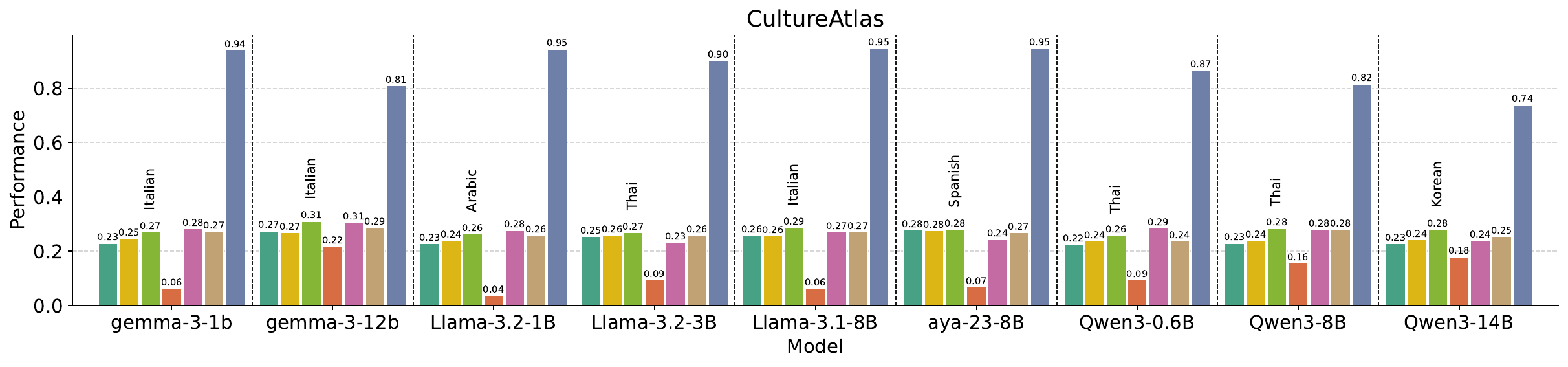}}
      \makebox[\textwidth][c]{\includegraphics[width=1\columnwidth]{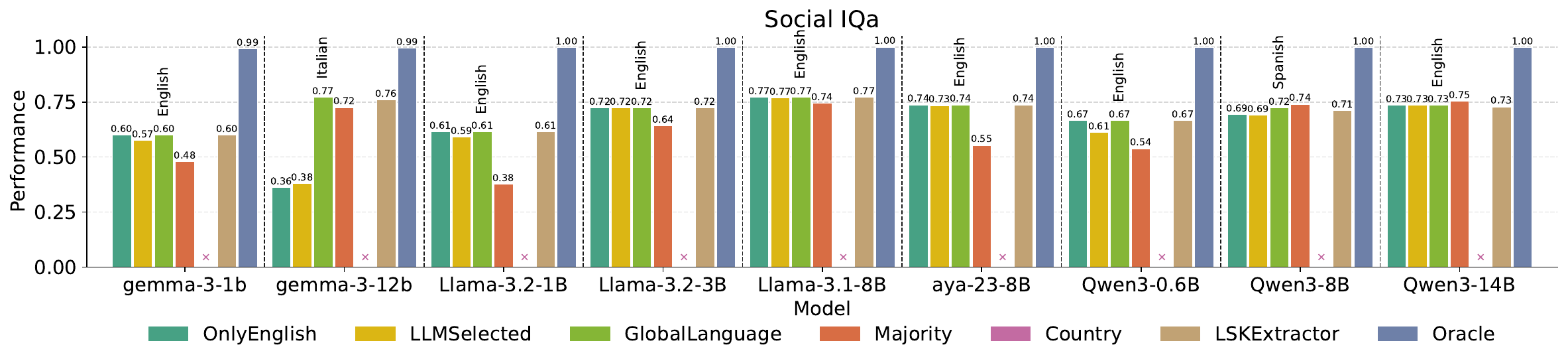}}
    \caption{The main results of measuring LSK -- we show the performance of our various baselines and LSK across the three datasets. Note: the text above the \textcolor{globallanguage}{Global Language} bar indicates the language that was selected the single language $L$ that achieves the highest overall accuracy. Also, Social IQa does not have results for \country because there is no such information in the dataset.}
    \label{fig: main_wr}
\end{figure}

\paragraph{There is considerable evidence that LSK exists and that the language selection problem is important.} First, the \onlyenglish baseline underperforms relative to the other methods, suggesting that restricting it to a single language can limit an LLM's effectiveness. This indicates that language choice is not merely a surface-level variation but can meaningfully influence model performance across tasks. Second, in situations where the \globallanguage method uses a \textit{different} language different from English, it performs better, indicating that English is not always the most intuitive language to use. This may reflect biases in training data distribution or differences in how knowledge is encoded and accessed across languages. Third, the \oracle method significantly outperforms all other approaches, nearly reaching 100\% accuracy. This result is particularly revealing: it shows that the required knowledge is already present within language models, but is not consistently accessible when queries are presented 
only in English. In other words, the limitation is not always because of a lack of knowledge, but of knowledge retrieval conditioned on language. As noted earlier, we introduce \oracle as a benchmark to quantify and ultimately reduce LSK-related knowledge gaps, with the broader goal of enabling more robust and language-agnostic model behavior.


\begin{figure}[t]
    \centering
     \makebox[\textwidth][c]{\includegraphics[width=\columnwidth]{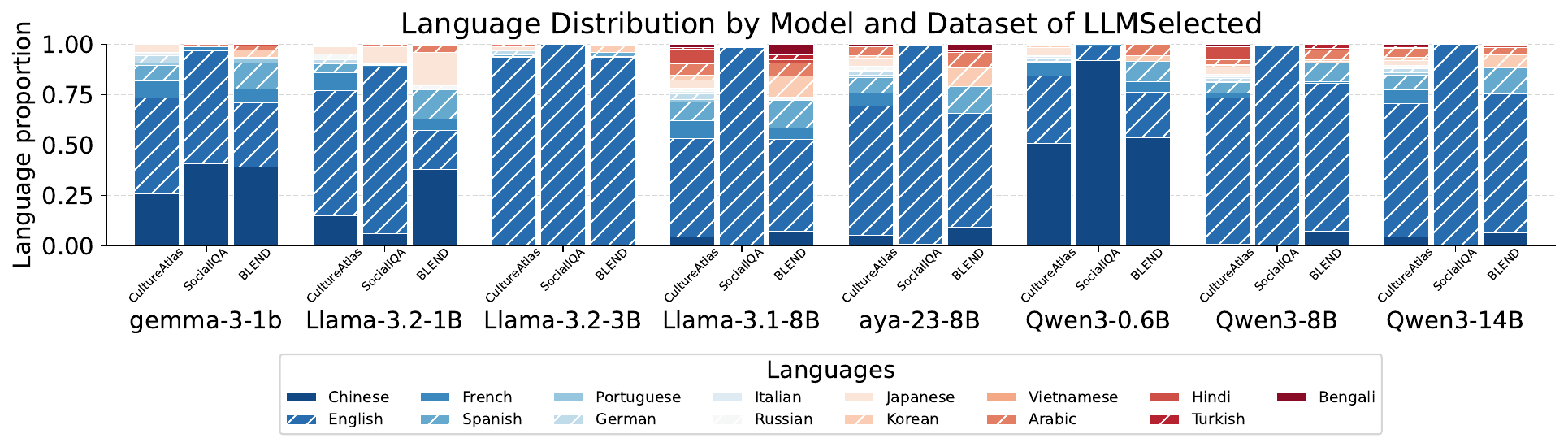}}
    \caption{Distribution of the languages selected by the \llmselected method, for each dataset and each model.}
    \label{fig: llm_selected_distribution}
\end{figure}

\begin{figure}[t]
    \centering
    \makebox[\textwidth][c]{\includegraphics[width=\columnwidth]{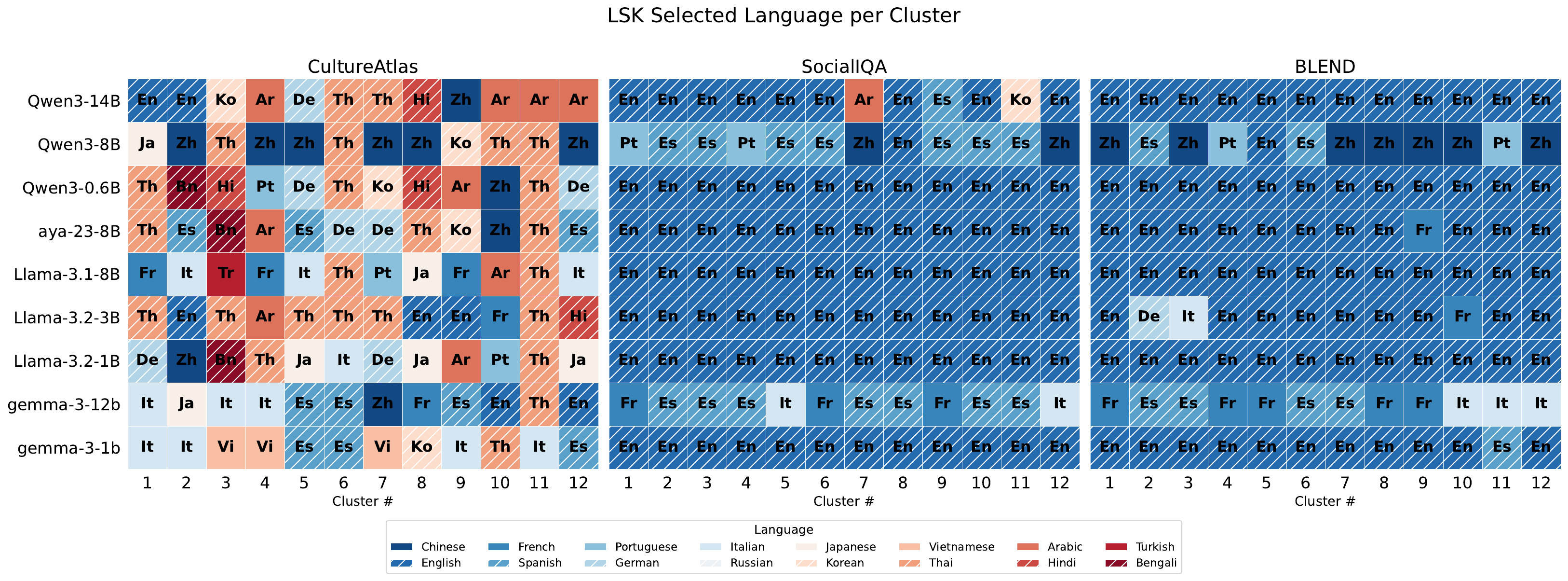}}
    \caption{A heatmap of the language selected by \lskextractor for each cluster, each model, and each dataset.}
    \label{fig: lsk_cluster_langs}
\end{figure}

\paragraph{Models might have internal language selection abilities.} This is evidenced by \llmselected's competitive performance. As shown in Figure \ref{fig: llm_selected_distribution}, models select a diverse range of languages for each dataset, including low-resource ones. However, the gap between \llmselected and \oracle suggests this internal capability could be more systematically leveraged in a more principled manner.

\paragraph{Strong methods such as \majority are not reliable.} Our initial intuition was that \majority would perform well on these benchmarks due to the large amount of multilingual context it provides. However, in many cases, it underperforms even the \onlyenglish baseline and simply increasing the number of languages does not necessarily improve performance, and may in fact introduce noise. This shows that knowledge is not represented equally in all languages, and that only certain languages may effectively surface the information required to answer a given question. This motivates the need to develop smarter, less expensive language selection methods to improve the overall performance of language models.

\paragraph{Intuitive language selection methods, such as \country are not sufficient.} 
The country-to-language mapping is based on the idea that models may store culturally or regionally specific knowledge more effectively in the corresponding local language. For example, one might expect an LLM to answer questions about Hindu culture better in Hindi, or questions about Turkish cuisine better in Turkish. If this assumption held, \country would be expected to outperform other baselines by a clear margin. However, this is not what we observe. The performance of \country suggests that such knowledge is not consistently best accessed through a single associated language. In addition, this approach is only applicable in settings where a clear country or regional label is available for each query, which limits its usefulness in more general scenarios. Overall, these results show that simple, intuition-driven mappings between topics and languages are not sufficient for reliable language selection. More robust methods are needed to identify the most effective language for each query in a data-driven way.

\renewcommand\tabularxcolumn[1]{m{#1}}          
\newcolumntype{Y}{>{\centering\arraybackslash}X} 

\begin{table}[t]

\centering
\scriptsize
\setlength{\tabcolsep}{4pt}
\renewcommand{\arraystretch}{0.95}

\begin{tabularx}{\textwidth}{|c|Y|Y|Y|}   
\toprule
\textbf{Cluster} & \textbf{BLEND Theme} & \textbf{CultureAtlas Theme} & \textbf{SocialIQA Theme} \\ \midrule
\midrule
1  & Regional specialties \& industries (livestock, agriculture, tourism) & Eastern/Central European countries (Ukraine, Serbia, Czech Rep., etc.) & Basic daily activities \& routine behaviors \\ \midrule
2  & Commercial hubs \& popular destinations & Western countries (France, Canada, Ireland, Sweden) & Social interactions \& interpersonal dynamics \\ \midrule
3  & Cultural celebrations \& traditional items (festivals, alcohol, ceremonies) & United States (exclusively) & Helping behaviors \& consideration for others \\ \midrule
4  & Economic centers \& basic needs (manufacturing, breakfast, commercial hubs) & Sub-Saharan African countries (Botswana, Niger, Ghana, etc.) & Goal-oriented actions \& planning \\ \midrule
5  & Specific regional activities (mining) \& entertainment & China (exclusively) & Personal interests \& character traits \\ \midrule
6  & Daily consumption \& rivalry (food, sports rivalries, quick meals) & Middle Eastern \& Mediterranean countries (Saudi Arabia, Greece, etc.) & Intimate relationships \& emotional connections \\ \midrule
7  & Sports achievements \& food origins (international success, global foods) & Southeast Asian \& Island nations (Philippines, Indonesia, Malaysia) & Authority, responsibility \& institutional roles \\ \midrule
8  & Family traditions \& regional preferences (meals, skiing, literature) & South Asian countries (India, Bangladesh, Nepal) & Complex social situations \& problem-solving \\ \midrule
9  & Cultural landmarks \& formal occasions (historic sites, weddings) & East Asian countries (Japan, with some Fiji) & Social dynamics \& behavioral expectations \\ \midrule
10 & Famous personalities (athletes, entrepreneurs) & Oceania (Australia, New Zealand, Papua New Guinea) & Material generosity \& preparation activities \\ \midrule
11 & Tourism \& technology hubs (attractions, sports, tech centers) & Southeast Asian countries (Thailand, Myanmar, Cambodia, Laos) & Professional care \& assistance behaviors \\ \midrule
12 & Competitive activities \& popular culture (sports teams, food) & Latin American \& Iberian countries (Mexico, Brazil, Spain, Peru) & Goal achievement \& recreational activities \\
\bottomrule
\end{tabularx}
\caption{Cluster themes for 12 clusters across datasets (related to the results in Figure \ref{fig: main_wr}).}
\label{tab: 12_cluster_analysis}
\end{table}

\begin{figure}[h]
    \centering
    \makebox[\textwidth][c]{\includegraphics[width=\columnwidth]{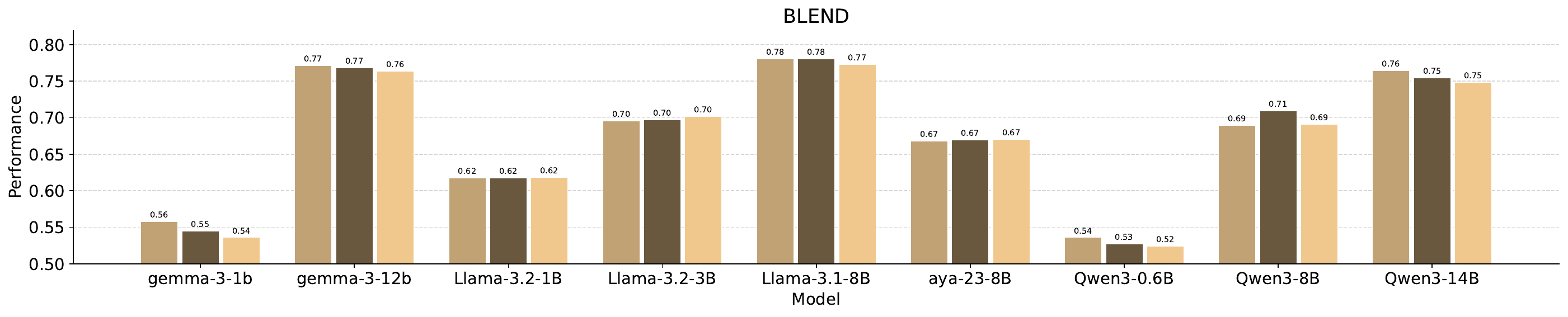}}
     \makebox[\textwidth][c]{\includegraphics[width=\columnwidth]{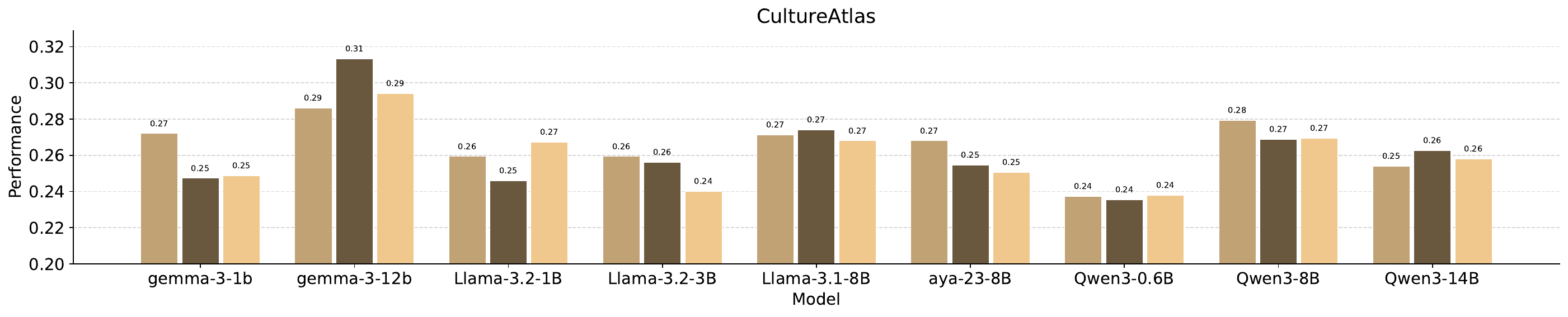}}
      \makebox[\textwidth][c]{\includegraphics[width=\columnwidth]{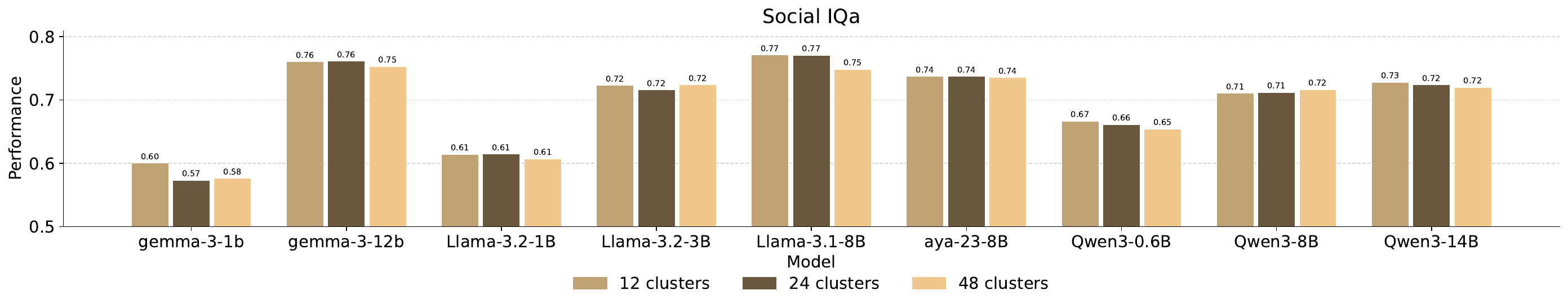}}
    \caption{Results on the effect of varying cluster size on the performance of \lskextractor.}
    \label{fig: cluster_analysis}
\end{figure}

\paragraph{There exists a semantic mapping of question to language.} For more clarity, we analyze the general topics of each \lskextractor cluster in Table \ref{tab: 12_cluster_analysis}. Alongside this, Figure \ref{fig: lsk_cluster_langs} plots heatmaps for the selected languages by \lskextractor. We see some intuitive trends: English is generally dominant because most of the training data for these models is in English. Furthermore, many models know the most about Southeast Asian countries (CultureAtlas, cluster 11) in Thai. Other trends are counter-intuitive/interesting: (1) Gemma models know about the most about China (CultureAtlas, cluster 5) and Middle Eastern countries (CultureAtlas, cluster 6) in Spanish, (2) large Qwen models perform best on authority and responsibility (Social IQA, cluster 7) in Arabic and Chinese, and (3) Cohere's Aya model knows the most about cultural landmarks (BLEND, cluster 9) in French. These patterns suggest that the relationship between topic and language is complex and effective language selection may depend on how knowledge is distributed in training data, rather than solely on real-world associations between topics and languages.

\lskextractor's performance shows that mapping question content to language is a promising approach. We use a cluster size of 12, though Figure \ref{fig: cluster_analysis} shows performance is mostly agnostic to cluster size (12, 24, 48). Future work can investigate more optimal content-to-language mappings.

\section{Conclusion}
In this paper, we explore the concept of Language Specific Knowledge (LSK)---languages contain specific knowledge not present in other languages. We also introduce the language selection problem. We empirically show evidence of LSK with seven different language selection methods. Broadly, we hope to encourage researchers to solve this language selection problem. In our work, we use signals such as model performance (\globallanguage, \majority), and semantic content (\country, \lskextractor), and verbalized model signals (\llmselected). Future work could work with lower-level, and more informative model signals such as confidence, gradient magnitude, hidden states, and attention maps. Since the \oracle method is achieving such high performance, we see that this is not a problem of epistemic uncertainty. This is a problem of how to elicit the right knowledge from these models. Hence, we pose our work as a benchmark, and encourage future researchers to improve the state-of-the-art on the language selection problem.



\clearpage
\bibliography{colm2026_conference}
\bibliographystyle{abbrvnat}

\newpage
\appendix

\section{CultureAtlas Reformatting} \label{app:cultureatlas_reformatting}
The CultureAtlas dataset consists of cultural claims associated with specific countries, each annotated as either true or false. Because this binary classification setting is relatively simple and the dataset is imbalanced toward false claims, we reformatted it into a multiple-choice question (MCQ) format. In the reformatted version, each question presents four answer choices pertaining to the same country: one true claim and three false claims. The model is then tasked with identifying the true claim, transforming the problem into a more nuanced and challenging task that requires reasoning across all options. An illustrative example of a reformatted question is shown in Figure~\ref{fig:cultureatlas_example}.

\begin{figure}[h]
  \centering
  \begin{tcolorbox}[
    colback=gray!5!white,
    colframe=black!75!black,
    title=An Example MCQ Generated from Culture Atlas,
    boxrule=0.3mm,
    width=\textwidth,
    arc=1.5mm,
    auto outer arc
  ]

  Question: What is true about Samoa? \\

Answer Choices: \\
A. There are several different kinds of possible group structures in Samoan culture. \\
B. Violent crime is limited, but increasing, and public perception associates this with returns of ethnic Tongans who have been raised overseas. \\
C. There is no social stigma on being in prison (although that may change now too), but then of course it also does not serve as a deterrent against crimes. \\
D. On all other social occasions, the taualuga is usually the last dance to be performed.\\

Ground Truth Answer: A.
  
  \end{tcolorbox}
  \caption{An example of a reformatted CultureAtlas question. The original binary (true/false) claims are transformed into a multiple-choice format with four options about the same country: one true claim and three false claims. The model is required to select the true claim.}
  \label{fig:cultureatlas_example}
\end{figure}

\section{Country to Language Mapping} \label{app:country_mapping}
For our Country Mapping baseline, we assign a language $\mathcal{l}_i \in \mathcal{L}$ to each country in the dataset. For each country, we select the most commonly spoken language in the corresponding region. If the most common language is not included in $\mathcal{L}$, we default to English. The mappings used in our experiments are summarized in Table~\ref{tab:country_language_mappings}.
\begin{table*}[h!]
\centering
\small
\begin{tabularx}{\textwidth}{p{1.8cm}p{1.5cm}X}
\toprule
\textbf{Dataset} & \textbf{Language} & \textbf{Countries} \\
\midrule
\multirow{5}{*}{\textbf{Blend}} 
 & \textbf{Arabic} & Algeria, Ethiopia \\
 & \textbf{Chinese} & China \\
 & \textbf{English} & Assam, Azerbaijan, Greece, Indonesia, Iran, Northern Nigeria, UK, US, West Java \\
 & \textbf{Korean} & North Korea, South Korea \\
 & \textbf{Spanish} & Mexico, Spain \\
\midrule
\multirow{15}{*}{\textbf{CultureAtlas}} 
 & \textbf{Arabic} & Algeria, Bahrain, Comoros, Egypt, Iraq, Jordan, Kuwait, Lebanon, Libya, Mauritania, Morocco, Oman, Qatar, Saudi Arabia, Sudan, Tunisia, United Arab Emirates, Yemen \\
 & \textbf{Bengali} & Bangladesh \\
 & \textbf{Chinese} & China \\
 & \textbf{English} & Afghanistan, Albania, Andorra, Antigua and Barbuda, Armenia, Australia, Azerbaijan, Bahamas, Barbados, Belarus, Belgium, Belize, Bhutan, Bosnia and Herzegovina, Botswana, Bulgaria, Burundi, Cambodia, Canada, Croatia, Cyprus, Czechia, Denmark, Dominica, Eritrea, Estonia, Eswatini, Ethiopia, Federated States of Micronesia, Fiji, Finland, Gambia, Georgia, Ghana, Greece, Grenada, Guyana, Haiti, Hungary, Iceland, Indonesia, Ireland, Islamic Republic of Iran, Israel, Jamaica, Kazakhstan, Kenya, Kiribati, Kyrgyzstan, Lao People's Democratic Republic, Latvia, Lesotho, Liberia, Lithuania, Luxembourg, Madagascar, Malawi, Malaysia, Maldives, Malta, Marshall Islands, Mauritius, Mongolia, Montenegro, Myanmar, Namibia, Nauru, Nepal, Netherlands, New Zealand, Nigeria, North Macedonia, Norway, Pakistan, Palau, Papua New Guinea, Philippines, Poland, Republic of Moldova, Romania, Rwanda, Saint Kitts and Nevis, Saint Lucia, Saint Vincent and the Grenadines, Samoa, Serbia, Seychelles, Sierra Leone, Singapore, Slovakia, Slovenia, Solomon Islands, Somalia, South Africa, South Sudan, Sri Lanka, Suriname, Sweden, Tajikistan, Timor-Leste, Tonga, Trinidad and Tobago, Turkmenistan, Tuvalu, Uganda, Ukraine, United Kingdom of Great Britain and Northern Ireland, United Republic of Tanzania, United States of America, Uzbekistan, Vanuatu, Zambia, Zimbabwe \\
 & \textbf{French} & Benin, Burkina Faso, Cameroon, Central African Republic, Chad, Congo, Côte d'Ivoire, Democratic Republic of the Congo, Djibouti, France, Gabon, Guinea, Monaco, Niger, Senegal, Togo \\
 & \textbf{German} & Austria, Germany, Liechtenstein, Switzerland \\
 & \textbf{Hindi} & India \\
 & \textbf{Italian} & Italy, San Marino \\
 & \textbf{Japanese} & Japan \\
 & \textbf{Korean} & Democratic People's Republic of Korea, Republic of Korea \\
 & \textbf{Portuguese} & Angola, Brazil, Guinea-Bissau, Mozambique, Portugal, São Tomé and Príncipe \\
 & \textbf{Russian} & Russian Federation \\
 & \textbf{Spanish} & Argentina, Bolivarian Republic of Venezuela, Chile, Colombia, Costa Rica, Cuba, Dominican Republic, Ecuador, El Salvador, Equatorial Guinea, Guatemala, Honduras, Mexico, Nicaragua, Panama, Paraguay, Peru, Plurinational State of Bolivia, Spain, Uruguay \\
 & \textbf{Thai} & Thailand \\
 & \textbf{Turkish} & Türkiye \\
 & \textbf{Vietnamese} & Viet Nam \\
\bottomrule
\end{tabularx}
\caption{Country-to-language mappings used for the Blend and CultureAtlas datasets. Each country is assigned its most commonly spoken language, defaulting to English if the language is not present in $\mathcal{L}$.}
\label{tab:country_language_mappings}
\end{table*}

\section{Model Prompts}
\label{app: model_prompts}
Figures \ref{prompt: with_reasoning_prompt_en} and \ref{prompt: with_reasoning_prompt_tr} contain the prompts to the language for during our evaluation, with reasoning, in English and Turkish (only two languages, to save space). Figure \ref{prompt: llm_selected_prompt} contains the prompt for the \llmselected. Figure \ref{prompt: translation} contains the prompt for translating the input dataset to various languages using GPT-4o-mini.

\begin{figure}[h]
  \centering
  \begin{tcolorbox}[
    colback=gray!5!white,
    colframe=black!75!black,
    title=Prompt to LLM in English,
    boxrule=0.3mm,
    width=\textwidth,
    arc=1.5mm,
    auto outer arc
  ]
  Question: \{input\_question\} \\
  Answer choices: \\
  A. \{choice\_one\} \\
  B. \{choice\_two\} \\
  C. \{choice\_three\} \\ 
  D. \{choice\_four\} \\

  Think about it in English, and then select one of the answer choices. Fill in the JSON below.
  \begin{verbatim}
{
  "reasoning_in_English": "<your reasoning steps in English>",
  "final_answer": "<output answer here>"
}
  \end{verbatim}
  \end{tcolorbox}
  \caption{Prompt to the language model to perform with reasoning, in English. Figure \ref{fig: main_wr} illustrates the results using this prompt. For BLEnD and Social IQa, the ``input\_question'' and ``choice\_x'' comes from the dataset. For CultureAtlas, because we modify the dataset ourselves to make it more difficult, the input question will always be ``Which is the following is true about \{country\}?''. Details of the CutlureAtlas modification are in Appendix \ref{app:cultureatlas_reformatting}.}
  \label{prompt: with_reasoning_prompt_en}
\end{figure}

\begin{figure}[h]
  \centering
  \begin{tcolorbox}[
    colback=gray!5!white,
    colframe=black!75!black,
    title=Prompt to LLM in Turkish,
    boxrule=0.3mm,
    width=\textwidth,
    arc=1.5mm,
    auto outer arc
  ]

  Soru: \{input\_question\} \\
  Cevap seçenekleri: \\
  A. \{choice\_one\} \\
  B. \{choice\_two\} \\
  C. \{choice\_three\} \\ 
  D. \{choice\_four\} \\

Türkçe olarak düşünün ve ardından cevap seçeneklerinden birini seçin. Aşağıdaki JSON'u doldurun.
  \begin{verbatim}
{
  "reasoning_in_Turkish": "<Türkçe akıl yürütme adımlarınız>",
  "final_answer": "<çıktı cevabı buraya>"
}
  \end{verbatim}
  \end{tcolorbox}
  \caption{Prompt to the language model to perform with reasoning, in Turkish. Figure \ref{fig: main_wr} illustrates the results using this prompt. For BLEnD and Social IQa, the ``input\_question'' and ``choice\_x'' comes from the dataset. For CultureAtlas, because we modify the dataset ourselves to make it more difficult, the input question will always be ``Which is the following is true about \{country\}?''. Details of the CutlureAtlas modification are in Appendix \ref{app:cultureatlas_reformatting}.}
  \label{prompt: with_reasoning_prompt_tr}
\end{figure}

\begin{figure}[h]
  \centering
  \begin{tcolorbox}[
    colback=gray!5!white,
    colframe=black!75!black,
    title=Prompt to LLM for selecting a language to best answer the question in (LLMSelected baseline),
    boxrule=0.3mm,
    width=\textwidth,
    arc=1.5mm,
    auto outer arc
  ]

An expert language is the language from the provided list that is most appropriate and informative for answering the given question (e.g., because the question is about a culture, region, or source where that language is dominant, or because that language has the richest knowledge base for the topic). \\

From the following languages: 

[Chinese, English, French, Spanish, Portuguese, German, Italian, Russian, Japanese, Korean, Vietnamese, Thai, Arabic, Hindi, Turkish, Bengali]

, determine which one is the best expert language for answering the question below. \\

Question: \{input\_question\} \\
Fill out your language expert in the below JSON format:
  \begin{verbatim}
{
 "expert_language": "<the expert language from the above list>"
}
  \end{verbatim}
  \end{tcolorbox}
  \caption{Prompt to the language model to select the language expert for a given question, i.e., \llmselected.}
  \label{prompt: llm_selected_prompt}
\end{figure}

\begin{figure}[h]
  \centering
  \begin{tcolorbox}[
    colback=gray!5!white,
    colframe=black!75!black,
    title=Dataset Translation Prompt to GPT-4o-mini,
    boxrule=0.3mm,
    width=\textwidth,
    arc=1.5mm,
    auto outer arc
  ]

  Translate ONLY the following question into \{language\}: "\{input\}".
  
  ONLY output the translation in the following JSON format:
  \begin{verbatim}
{
    "{language}_translation": <output the translated input 
        here>.
}
  \end{verbatim}
  \end{tcolorbox}
  \caption{Prompt to GPT-4o-mini to translate the datasets into one of the 16 languages we chose for our experimentation. As input, the translation ``language'' and the text to translate (``input'') is provided.}
  \label{prompt: translation}
\end{figure}

\section{Output Analysis}
\label{app: output_analysis}
In order to verify whether the models are outputting responses that align to the language they are supposed to reason in, we run a language classification model (specifically, \texttt{qanastek/51-languages-classifier} -- we choose this for its good performance, and because it covers the language set we choose for our experimentation) and calculate the percentage of samples that follow the intended reasoning language. For BLEnD, CultureAtlas, and Social IQa, respectively, we see average accuracies of 96.97\%, 97.73\% and 97.93\%, across all models and languages. This indicates that models generally are very good at following instructions to think in a certain language, and further strengthens the claims we make in our paper.

\clearpage
\section{Experimental Specifications}
\label{app: resources}
We run our inference on NVIDIA A40 GPUs. For the the 1B, 3B, 8B models, we used a single A40 GPU, while the 12B and 14B required two A40 GPUs. Inference takes around 30-60 minutes per language. Clustering is computationally inexpensive and can be done on a single A40 GPU. 

\section{Licenses}
\label{app: licenses}
Our \href{https://anonymous.4open.science/r/LSKExtractor-272F/}{code} is released publicly under the Apache-2.0 License. CultureAtlas \citep{cultureatlas} is released under the MIT License; BLEnD \citep{blend} under the CC-by-SA-4.0 License; SocialIQa \citep{socialiqa} is not under explicit license, however it is publicly available on Huggingface, and we do not use it for commercial purposes. All models are under their proprietary licenses from the corresponding companies.

\section{Use of Large Language Models}
Other than being used as part of the experiments conducted in this work, LLMs were used solely as a writing assistance tool in preparing this paper submission. Their role was limited to polishing language, improving clarity, and reducing redundancy. The prompt used for this purpose was similar to "Please revise the writing of this, making sure to remove any grammatical mistakes." All research ideas, experimental designs, analyses, and claims presented in the paper are entirely the original work of the authors. No part of the conceptual, methodological, or empirical contributions relies on or originates from LLM outputs.
\end{document}